\documentclass[graybox]{svmult}

\usepackage{mathptmx}       
\usepackage{helvet}         
\usepackage{courier}        
\usepackage{type1cm}        
\usepackage{makeidx}         
\usepackage{graphicx}        
\usepackage{multicol}        
\usepackage[bottom]{footmisc}
\usepackage[square,numbers]{natbib}

\makeindex             

\begin{document}

\title*{Bangla Handwritten Digit Recognition and Generation}
\author{Md Fahim Sikder}
\institute{Md. Fahim Sikder \at Department of Computer Science \& Engineering, Jahangirnagar University, Savar, Bangladesh, \email{fahimsikder01@gmail.com}}
%
%
\maketitle

\abstract*{Handwritten digit or numeral recognition is one of the classical issues in the area of pattern recognition and has seen tremendous advancement because of the recent wide availability of computing resources. Plentiful works have already done on English, Arabic, Chinese, Japanese handwritten script. Some work on Bangla also have been done but there is space for development. From that angle, in this paper, an architecture has been implemented which achieved the validation accuracy of 99.44\% on BHAND dataset and outperforms Alexnet and Inception V3 architecture. Beside digit recognition, digit generation is another field which has recently caught the attention of the researchers though not many works have been done in this field especially on Bangla. In this paper, a Semi-Supervised Generative Adversarial Network or SGAN has been applied to generate Bangla handwritten numerals and it successfully generated Bangla digits.}

\abstract{Handwritten digit or numeral recognition is one of the classical issues in the area of pattern recognition and has seen tremendous advancement because of the recent wide availability of computing resources. Plentiful works have already done on English, Arabic, Chinese, Japanese handwritten script. Some work on Bangla also have been done but there is space for development. From that angle, in this paper, an architecture has been implemented which achieved the validation accuracy of 99.44\% on BHAND dataset and outperforms Alexnet and Inception V3 architecture. Beside digit recognition, digit generation is another field which has recently caught the attention of the researchers though not many works have been done in this field especially on Bangla. In this paper, a Semi-Supervised Generative Adversarial Network or SGAN has been applied to generate Bangla handwritten numerals and it successfully generated Bangla digits.}

\section{Introduction}
\label{intro}

Recognizing handwritten numerals is one of the emerging problems in the sector of computer vision. Automation of the banking system, postal services, form processing are the practical example of handwritten character recognition \cite{pal2009lexicon,pal2012multi,yacoubi2001handwritten,bunke2004offline,madhvanath1995reading,srihari1995name,bhowmik2018off}. A lot of work already has been done with great accuracy in the recognition of English handwritten digits \cite{bengio2007greedy,lecun1995comparison}. Researchers used support vector machine, histogram of gradient oriented, neural network etc algorithm to solve these problems. Recently, a lot of focus has been drawn to the neural network architecture due to the wide availability of high-performance computing systems \cite{abir2019bangla}. ANNs are computing system which is influenced by the organic neural network. Convolutional Neural Network is one of the architectures of neural network which makes it easy to recognize image with great accuracy. Besides English, a lot of work also done in Arabic, Chinese, Japanese and Roman scripts \cite{broumandnia2008persian,el2015word,dehghan2001handwritten,liu2002lexicon,su2013chinese,srihari2007offline,koerich2005recognition,bunke2003recognition,bozinovic1989off}. But in the case of Bangla, not many works have been done and there is a chance for improvement.

On the other hand, generating images is another outstanding image processing field recently caught the attention of researchers. Image generation can be used in art creation, fraudulent detection also can be applied in law enforcement. Generative Adversarial Network or GAN, another architecture of neural network is been used to generate the image. Researchers also applied GAN to generate MNIST dataset but not much work has been done in other datasets.
To mend this research gap on Bangla, we have implemented an architecture which recognizes Bangla handwritten digits at 99.44\% accuracy using BHAND dataset which contains 70000 images of Bangla handwritten digits which are collected from 1750 persons. At the same time, we have implemented a semi-supervised generative adversarial network or SGAN to generate Bangla digits. The paper is arranged as follows: Section \ref{relate} reviews the relevant works, Section \ref{proposed} describes the proposed solution, Section \ref{experiment} describes the result and lastly, Section \ref{conc} concludes the paper.


\section{Related Works}
\label{relate}

A lot of research works have been done on Bangla handwritten digit recognition using SVM \cite{bhowmik2009svm}, HOG \cite{bhattacharya2009handwritten} etc. Recently loads of attention is being given on deep learning because of easy access to GPU (graphics processing unit). Using multilayer convolutional layer, pooling layer increases the performance of accuracy. Some of the legendary deep learning based architecture such as Alexnet \cite{krizhevsky2012imagenet}, LeNet \cite{lecun1990handwritten}, Inception V3 \cite{szegedy2015going} took the accuracy of image recognition to the next level. MNIST recognition \cite{lecun1989backpropagation}, CIFAR-10 database recognition \cite{krizhevsky2012imagenet} are some example of that architecture. For Bangla handwritten recognition numerous work has been done. But initially, it was troublesome for the researcher because of the limitation of a dataset \cite{akhand2015bangla}. But now some great datasets are available for Bangla digit recognition. A deep belief network is being introduced where the author first used unsupervised feature learning then it's followed by a supervised fine-tuning \cite{sazal2014bangla}. In \cite{saadat2016b}, the author removed overfitting problem and has an error rate of 1.22\%.

Besides digit recognition, few works have been done on digit generation. Researchers used different kinds of generative adversarial networks (GAN) to generate digits or characters. Auxiliary Classifier GAN \cite{odena2016conditional}, Bidirectional GAN \cite{donahue2016adversarial}, Deep Convolutional GAN \cite{radford2015unsupervised}, Semi-Supervised GAN \cite{odena2016semi} were used on MNIST dataset to generate digits.  

\section{Proposed Work}
\label{proposed}
In this work, we have proposed a architecture for digit recognition which outperforms Alexnet \cite{krizhevsky2012imagenet} and Inception V3 \cite{szegedy2015going} model at validation accuracy and error on the BHAND \cite{saadat2016b} dataset. Also, we have implemented Semi-Supervised Generative Adversarial Network (SGAN) for digit generation for the same dataset.

\subsection{Dataset Description}
For recognition and generation, BHAND dataset has been used which contains 70000 handwritten Bangla digits. This is one of the biggest datasets of handwritten Bangla digits. This dataset is divided into three sets: Training set (50000), Testing set (10000) and Validation set (10000). These 70000 data is collected from 1750 persons. The images are gray-scale and the dimension is $32*32$.

\subsection{Network Architecture}
\label{sub:network}

For recognizing handwritten digit, we have proposed an architecture which consists several convolutional layers, pooling layers, normalization layers, and dense or fully connected layers. In the first convolutional layer, we took the $32 * 32$ images as input from the dataset. As mentioned earlier the images are grayscale, so it has $1$ channel. In this layer, we have taken $32$ filter which has the filter size of $2*2$. 
\begin{figure}
	\centering
	\includegraphics[width=.75\textwidth]{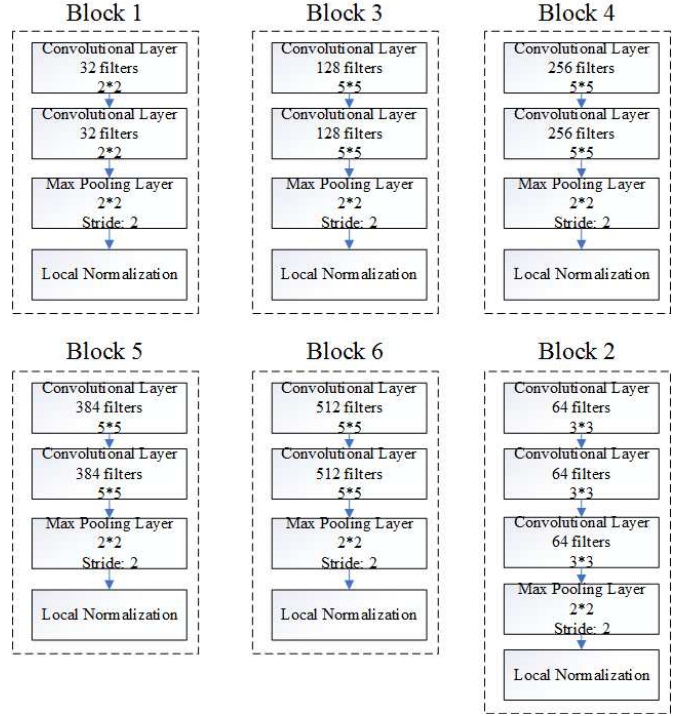}
	\caption{Blocks of the architecture}
	\label{fig:block}       
\end{figure}
The output of this layer then goes into a second convolutional layer which also has $32$ filters and the size of those filters is $2*2$. Then the outcome of the second convolutional layer feed into max pooling layer which has the filter size of $2*2$ and the stride size is $2$. This outcome then goes into a normalization layer. These convolutional layers, pooling layer and normalization layer, together we named it $block$. In a single block the number of these layers could vary. The second block is composed of three convolutional layers, one max pooling layer, and another normalization layer. The amount of filters in the second block’s convolutional layers are $64$ and the filter size is $3*3$. This max pooling layer has also $2*2$ filter size and stride of $2$. Then the third to sixth block consists of two convolutional layers, one pooling layer and one normalization layer. Third block’s convolutional layer has $128$ filters and the size of the filters is $5*5$, fourth block’s convolutional layer has $256$ filters which has $5*5$ filter size, fifth block’s convolutional layer has $384$ filters, sixth block’s convolutional layer has $512$ filters and their filter size is $5*5$. And all the blocks have the same pooling layer architecture. It has $2*2$ filter size and stride size $2$. Figure \ref{fig:block} shows the blocks used in this architecture.

The outcome of the sixth block then feed into a fully connected layer which has $1024$ units then we drop the $50\%$ of the neuron for avoiding overfitting then the output is fed on the second fully connected layer which has $5120$ units. Here we also drop the $50\%$ of the neuron. Till now every layer used $relu$ activation function. The following equation \cite{WinNT} is how $relu$ works. 

\[R(z) = max(0,z)\]

Now the output is then fed into the last fully connected layer which has $10$ units because we have $10$ class as output and here we have used $softmax$ activation function. The following equation \cite{WinNT} is how $softmax$ works.

\[s(z)_j = \frac{e^{z_j}}{\sum_{k=1}^{K}e^{z_k}}\] 

\begin{figure}
	\centering
	\includegraphics[width=.75\textwidth]{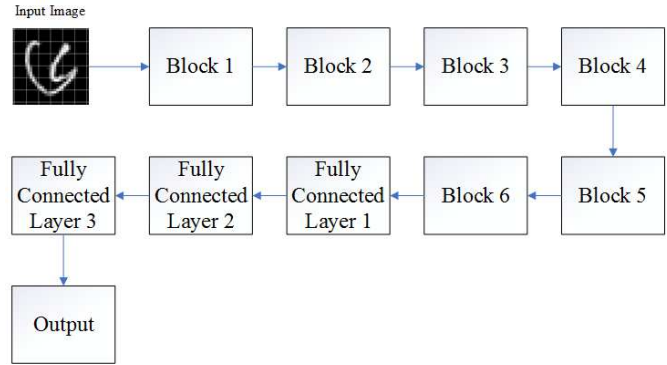}
	\caption{Our architecture for digit recognition}
	\label{fig:recog}       
\end{figure}

The complete architecture of the recognizing part is shown in figure \ref{fig:recog}.

\begin{figure}
	\centering
	\includegraphics[width=.75\textwidth]{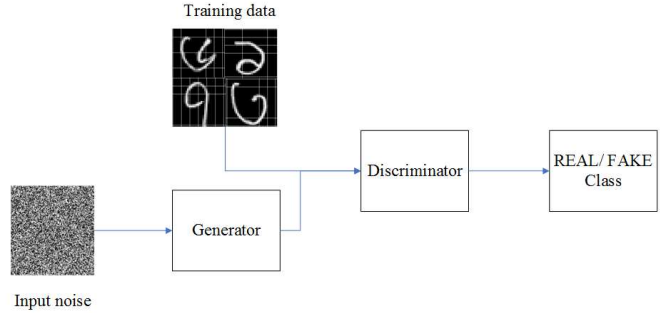}
	\caption{Architecture of SGAN}
	\label{fig:sgan}       
\end{figure}

Now for the digit generation part, here Semi-Supervised Generative Adversarial Network (SGAN) \cite{odena2016semi} is used for this task. Here we have a generator and discriminator. We took random noise as input, then the noise goes to the generator, at the same time we took a sample from training dataset. The generator attempts to forge the sample from training dataset and both the real and fake data goes to the discriminator then the discriminator attempts to distinguish between the genuine and the fabricated one. Usually, in GAN we train generator and discriminator concurrently and after training, we could discard discriminator because its only used for training the generator. In SGAN we alter the discriminator into a classifier and we discard the generator after the training. Here generator is used to aid the discriminator during training. Figure \ref{fig:sgan} shows the complete architecture of the SGAN.

In the generator, first, we took a random vector as input then we reshape it and then batch normalize it. Then we $upsample$ the output. After that, we took a convolutional layer and pass the output through it. The convolutional layer has $128$ filters and the filter size is $3*3$ also we used the $same$ padding. We again use batch normalize and upsample in it. After that, we use another convolutional layer which has the same filter size and padding but it has only 64 filters and we again batch normalize it. The last two convolutional layers used $relu$ activation function. Now the output is passed through the last convolutional layer which has one filter and the filter size and padding are like the same as others and it used $tanh$ activation function. Now for the discriminator part, it is a multiclass classifier. We have used four convolutional layers. First convolutional layer takes $32*32$ images and it has $32$ filters which have the size of $3*3$ also the strides of 2 to reduce the dimension of the feature vectors. Here we have used $Leaky Rectified Linear Unit$ activation functions.
Then we drop 25\% of neurons for avoiding overfitting. Then the output goes to the next convolutional layers which have 64 filters and the size and strides are same as the last one. Then again, we drop 25\% of neuron and use batch normalization. In the third and fourth convolutional layer, the filter size is the same but has 128 and 256 filters respectively. Then we flatten the output. In the end, we used two dense or fully connected layers. The last layer takes $N+1$ units because discriminator could generate $N+1$ outputs because of the fake label. Here is $N$ is the number of total class and we used $softmax$ activation function. We used $binary-crossentropy$ loss function and $Adam$ optimizer.

\section{Experimental Analysis \& Result}
\label{experiment}

We have implemented our architecture using BHAND dataset which has $50000$ training image, $10000$ testing image and $10000$ validating images of handwritten Bangla numerals. It has $32*32$ image dimension and the number of the channel was $1$. For recognizing the digit, we have also applied this dataset in popular alexnet and inception v3 model. We have run a total of $19550$ steps in the training and achieved $99.44\%$ validation accuracy. We have used $rmsprop$ optimizer and $categorical-crossentropy$ as loss function. The learning rate in our architecture was $0.001$. A detailed analysis of our experiments is shown in table \ref{tab:comp}.

\begin{table}[h!]
	\centering
	\caption{Comparison of our model with others for recognizing digit}
	\label{tab:comp}       
	\begin{tabular}{llll}
		\hline\noalign{\smallskip}
		Model Name & Steps & Validation Accuracy & Validation Error \\
		\noalign{\smallskip}\hline\noalign{\smallskip}
		Alexnet & 19550 & 97.74\% & 0.1032 \\
		Inception V3 & 19550 & 98.13\% & 0.07203 \\
		Our Model & 19550 & \textbf{99.44\%} & \textbf{0.04524} \\
		\noalign{\smallskip}\hline
	\end{tabular}
\end{table}

\begin{figure}
	\centering
	\includegraphics[width=.75\textwidth]{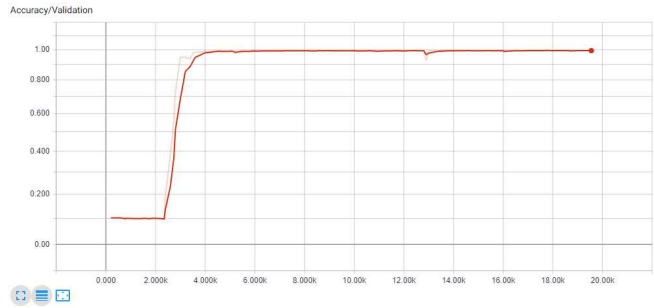}
	\caption{Validation Accuracy of our model for recognizing digit}
	\label{fig:vald}       
\end{figure}

\begin{figure}
	\centering
	\includegraphics[width=.75\textwidth]{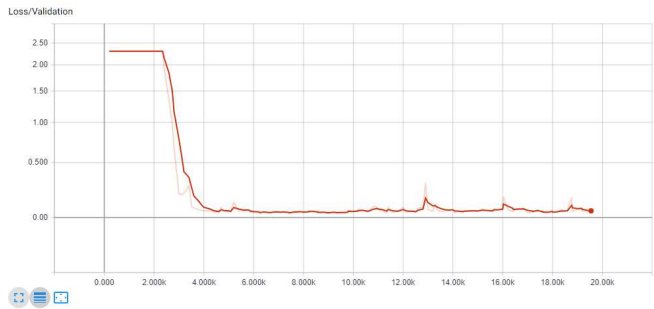}
	\caption{Validation Error of our model for recognizing digit}
	\label{fig:valderr}       
\end{figure}

The validation accuracy and the validation error of our model is shown respectively in figure \ref{fig:vald} and \ref{fig:valderr}.

For generating the Bangla handwritten image we also used the same dataset. For generating an image, we have used the Semi-Supervised Generative Adversarial Network (SGAN). Here we have built our model using generator and discriminator. Generator took a random vector as input. On the other hand from the real train dataset, an image goes to the discriminator. Generator tries to fool the discriminator by mimicking the real image. Then the discriminator discriminates the real and forges image. For our generator, we have used a combination of a fully connected layer, convolutional
\begin{figure}
	\centering
	\includegraphics[width=.95\textwidth]{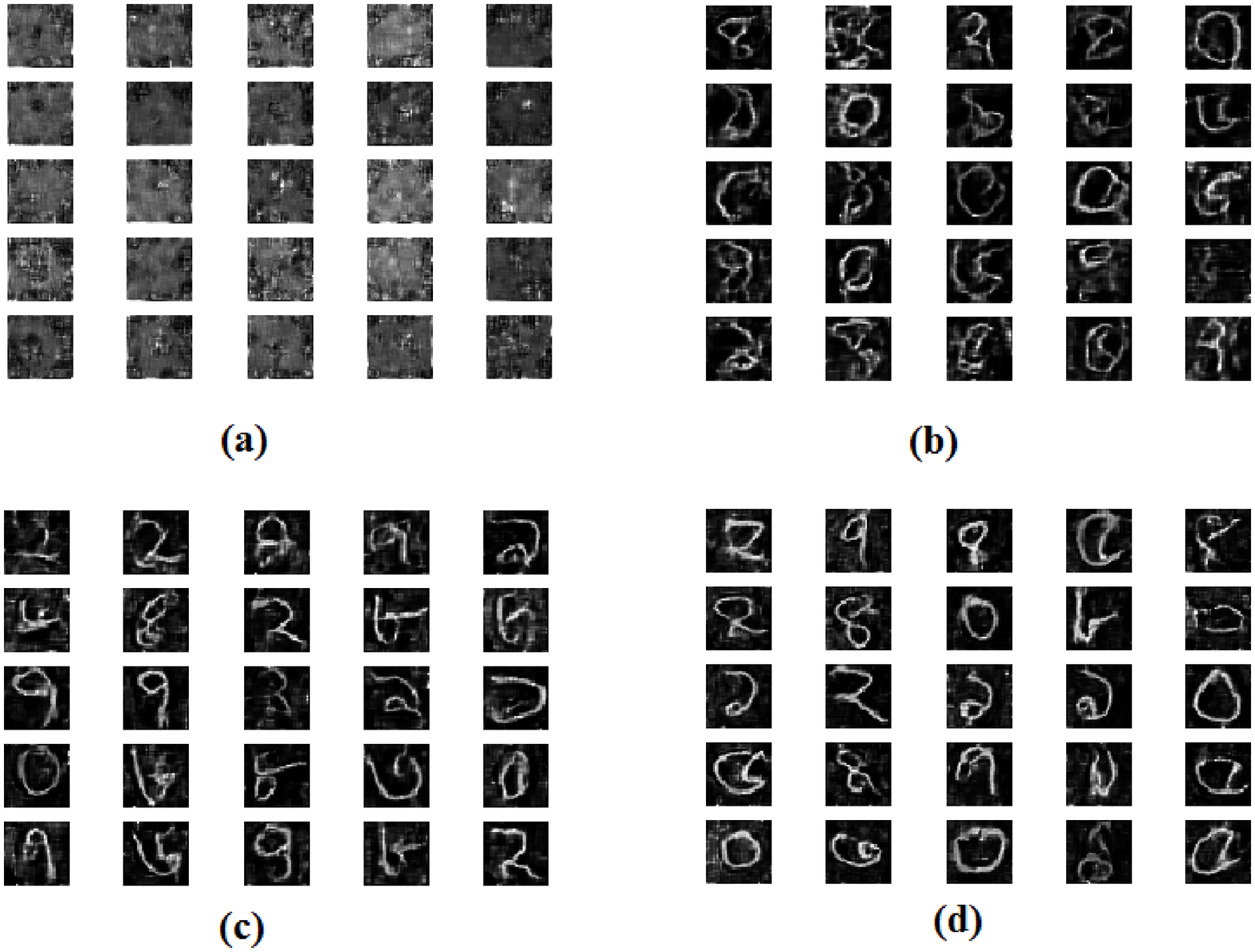}
	\caption{Output of our generation model at step 0, 100000, 200000 and 300000}
	\label{fig:comb}       
\end{figure}
layer. Also, we need to normalize and upsample our data. For the discriminator, it also has a series of a convolutional layer and fully connected layer. Discriminator took the image as input to the input dimension is $32*32$. It used two loss function: $binary-crossentropy$ and $categorical-crossentropy$ whereas generator used $binary-crossentropy$. Here we have used Adam optimizer 
\begin{figure}
	\centering
	\includegraphics[width=.75\textwidth]{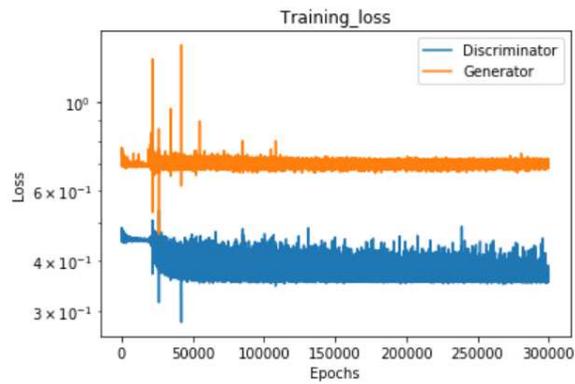}
	\caption{Training Loss of our model for digit generation}
	\label{fig:genres}       
\end{figure}
where the learning rate is $0.002$. We have also reshaped our data to $-1$ to $1$ because of the usage of $sigmoid$ and $tanh$ activation function. After $300000$ steps of training, we have got $0.368$ loss of discriminator and $0.694$ generator loss. From figure \ref{fig:comb} we can see the output of our SGAN. The first image (a) is from $0$ step, the second image (b) is after $100000$ and (c) and (d) image are after respectively $200000$ and $300000$ steps. The training loss is shown in figure \ref{fig:genres}.

\section{Conclusion}
\label{conc}

Loads of work have been done in the area of handwritten numeral recognition but still, there is an opportunity to improve and only a few works has been done in the area of digit generation. From that motivation, in this paper, we have proposed a architecture for recognizing Bangla handwritten digits which outperforms popular alexnet and inception v3 architecture using BHAND dataset. By adding a more convolutional layer and hyperparameter tuning could result in a better performance. Also, we have implemented the Semi-Supervised Generative Adversarial Network (SGAN) using the same dataset and successfully generate Bangla digits. In the future, we will try to reduce the discriminator's training loss on SGAN.

\section*{Acknowledgment}
\label{ack}

The author is grateful to the anonymous reviewers for their comments that improved the quality of this paper, also thankful to Md. Rokonuzzaman Sir from ISTT and Umme Habiba Islam for their support and help.

\bibliographystyle{spbasic}      
\bibliography{generate}   

\end{document}